\documentclass[acmtog,screen]{acmart}

\AtBeginDocument{%
  \providecommand\BibTeX{{%
    \normalfont B\kern-0.5em{\scshape i\kern-0.25em b}\kern-0.8em\TeX}}}

\copyrightyear{2022}
\acmYear{2022}
\setcopyright{licensedothergov}\acmConference[SIGGRAPH '22 Conference Proceedings]{Special Interest Group on Computer Graphics and Interactive Techniques Conference Proceedings}{August 7--11, 2022}{Vancouver, BC, Canada}
\acmBooktitle{Special Interest Group on Computer Graphics and Interactive Techniques Conference Proceedings (SIGGRAPH '22 Conference Proceedings), August 7--11, 2022, Vancouver, BC, Canada}
\acmPrice{15.00}
\acmDOI{10.1145/3528233.3530738}
\acmISBN{978-1-4503-9337-9/22/08}

\newcommand{\bc}{\mathbf{c}}
\newcommand{\bD}{\mathbf{D}}

\newcommand{\bF}{\mathbf{F}} %
\newcommand{\bG}{\mathbf{G}}

\newcommand{\bP}{\mathbf{P}}

\newcommand{\bw}{\mathbf{w}}
\newcommand{\bx}{\mathbf{x}}

\newcommand{\bz}{\mathbf{z}}

\newcommand{\nE}{\mathbb{E}}

\newcommand{\cL}{\mathcal{L}}

\newcommand{\figref}[1]{Fig.~\ref{#1}}

\newcommand{\tabref}[1]{Table~\ref{#1}}

\makeatletter
\DeclareRobustCommand\onedot{\futurelet\@let@token\@onedot}
\def\@onedot{\ifx\@let@token.\else.\null\fi\xspace}

\makeatother

\newcommand{\boldparagraph}[1]{\vspace{0.2cm}\noindent{\bf #1} }

\definecolor{darkgreen}{rgb}{0,0.7,0}
\definecolor{darkblue}{RGB}{31,119,180}
\definecolor{darkred}{RGB}{214,39,40}

\usepackage{appendix}

\providecommand{\impath}[1]{}
\providecommand{\impatha}[1]{}
\providecommand{\impathb}[1]{}
\providecommand{\impathc}[1]{}
\providecommand{\impathd}[1]{}
\providecommand{\impathe}[1]{}

\newcommand{\teaser}{
\begin{teaserfigure}
 \includegraphics[width=1.00\textwidth, trim=0em 0em 0em 0em, clip]{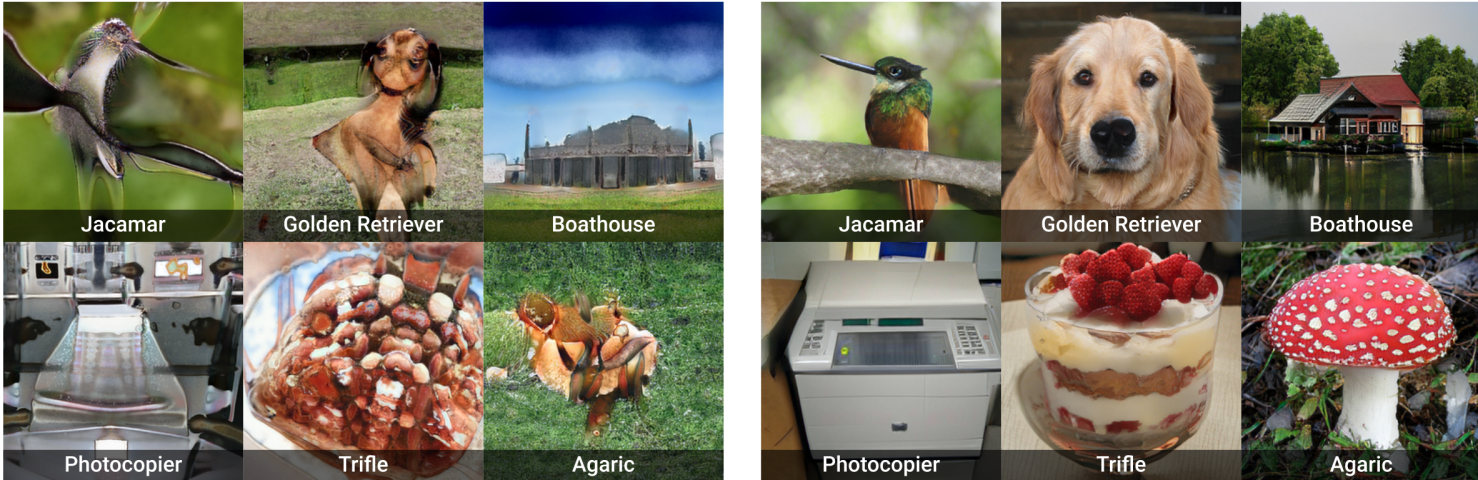}
  \caption{ 
  Class-conditional samples generated by StyleGAN3 (left) and StyleGAN-XL (right) trained on ImageNet at resolution $256^2$.
  }
  \label{fig:teaser}
\end{teaserfigure}
}

\newcommand{\system}{
\begin{figure*}
  \includegraphics[width=\textwidth]{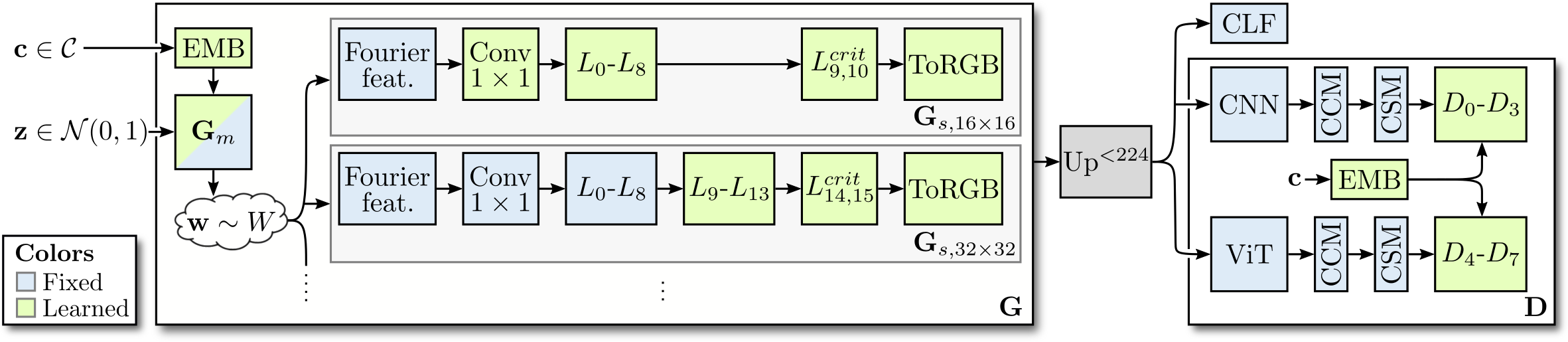}
  \caption{\textbf{Training StyleGAN-XL}.
  We feed a latent code $\bz$ and class label $\bc$ to the pretrained embedding and the mapping network $\bG_m$ to generate style codes $\bw$. The codes modulate the convolutions of the synthesis network $\bG_s$. During training, we gradually add layers to double the output resolution for each stage of the progressive growing schedule. We only train the latest layers while keeping the others fixed. $\bG_m$ is only trained for the initial $16^2$ stage and remains fixed for the higher-resolution stages. The synthesized image is upsampled when smaller than $224^2$ and passed through a CNN and a ViT and respective feature mixing blocks (CCM+CSM). At higher resolutions, the CNN receives the unaltered image while the ViT receives a downsampled input to keep memory requirements low but still utilize its global feedback. Finally, we apply eight independent discriminators on the resulting multi-scale feature maps. The image is also fed to classifier CLF for classifier guidance.
  }
  \label{fig:system}
\end{figure*}
}

\newcommand{\highres}{
\begin{figure*}[!p]
  \includegraphics[width=\textwidth]{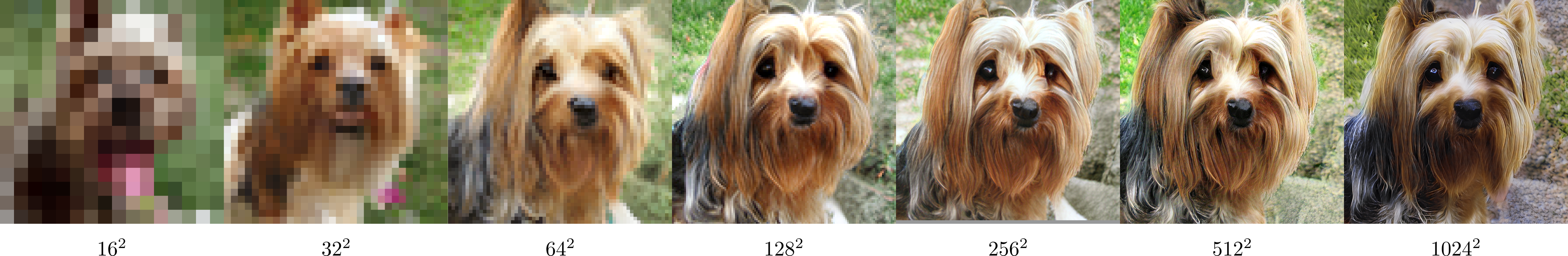}
  \caption{\textbf{Samples at Different Resolutions Using the Same $\bw$}. The samples are generated by the models obtained during progressive growing. We upsample all images to $1024^2$ using nearest-neighbor interpolation for visualization purposes. Zooming in is recommended.
  }
  \label{fig:highres}
\end{figure*}
}

\newcommand{\inversion}{
\begin{figure*}[!p]
  \includegraphics[width=\textwidth]{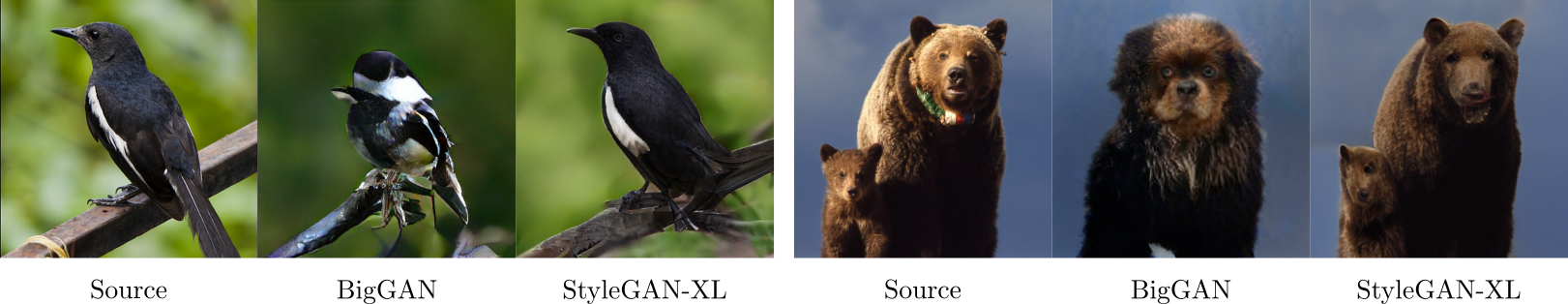}
  \caption{\textbf{Inversion of a Given Source Image}.
  For BigGAN, we invert to its latent space $\bz$, for StyleGAN-XL we invert to style codes $\bw$.
  }
  \label{fig:inversion}
\end{figure*}
}

\newcommand{\interpolations}{
\begin{figure*}[!p]
  \includegraphics[width=\textwidth]{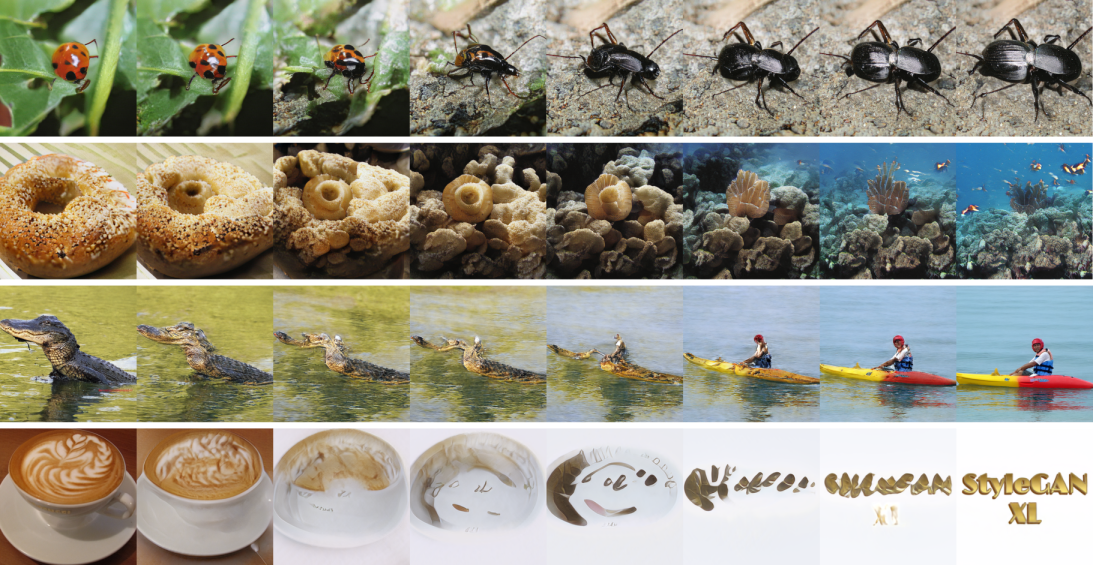}
    \caption{\textbf{Interpolations}. StyleGAN-XL generates smooth interpolations between samples of different classes (Row 1 \& Row 2). PTI allows inverting to the latent space with low distortion (outermost image, Row 3 \& Row 4), and consistently embeds out-of-domain inputs, such as the one on the bottom right.
  }
  \label{fig:interpolations}
\end{figure*}
}

\newcommand{\editing}{
\begin{figure*}[!p]
  \includegraphics[width=\textwidth]{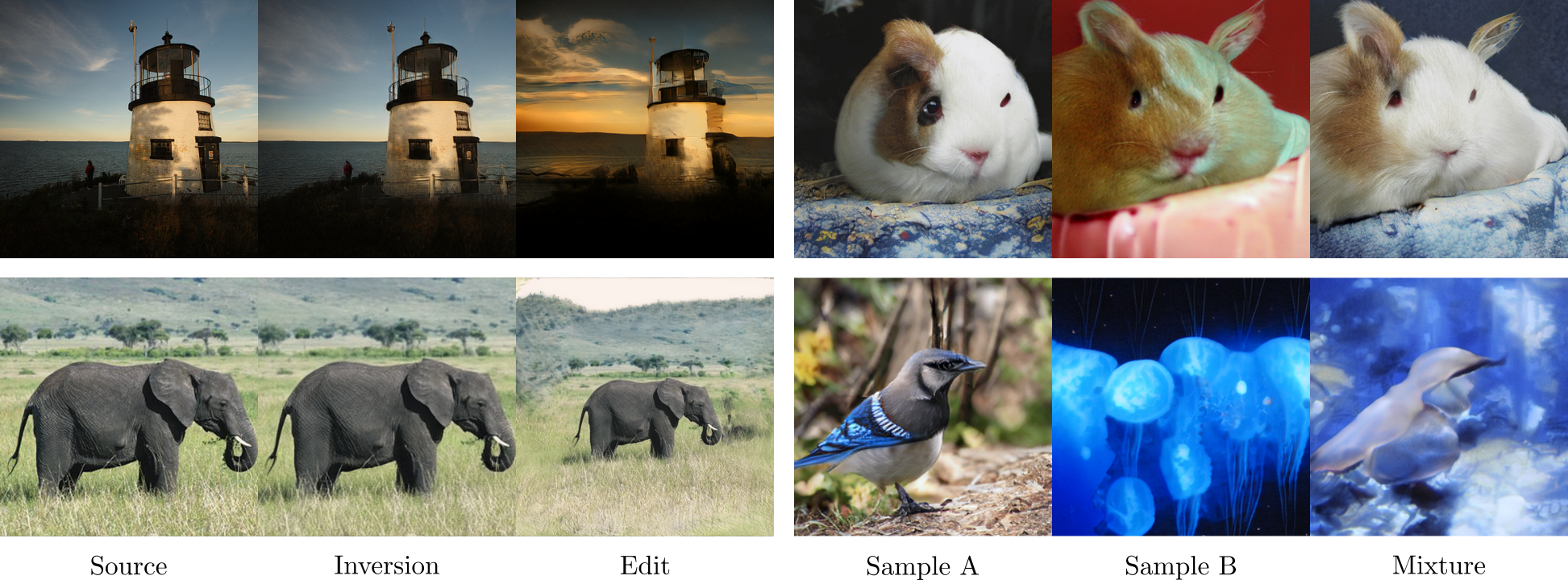}
\caption{\textbf{Image Editing and Style Mixing}. 
Left: First, a given image is inverted via PTI~\cite{Roich2021ARXIV}.
Right: Given two images, we can mix their styles. This methods works for samples of the same or similar classes, and to a certain extent, for distant classes. For this experiment, we utilize random samples instead of inversions.
  }
  \label{fig:editing}
\end{figure*}
}

\newcommand{\layerspecs}{
\begin{figure*}[!p]
  \includegraphics[height=0.9\textheight]{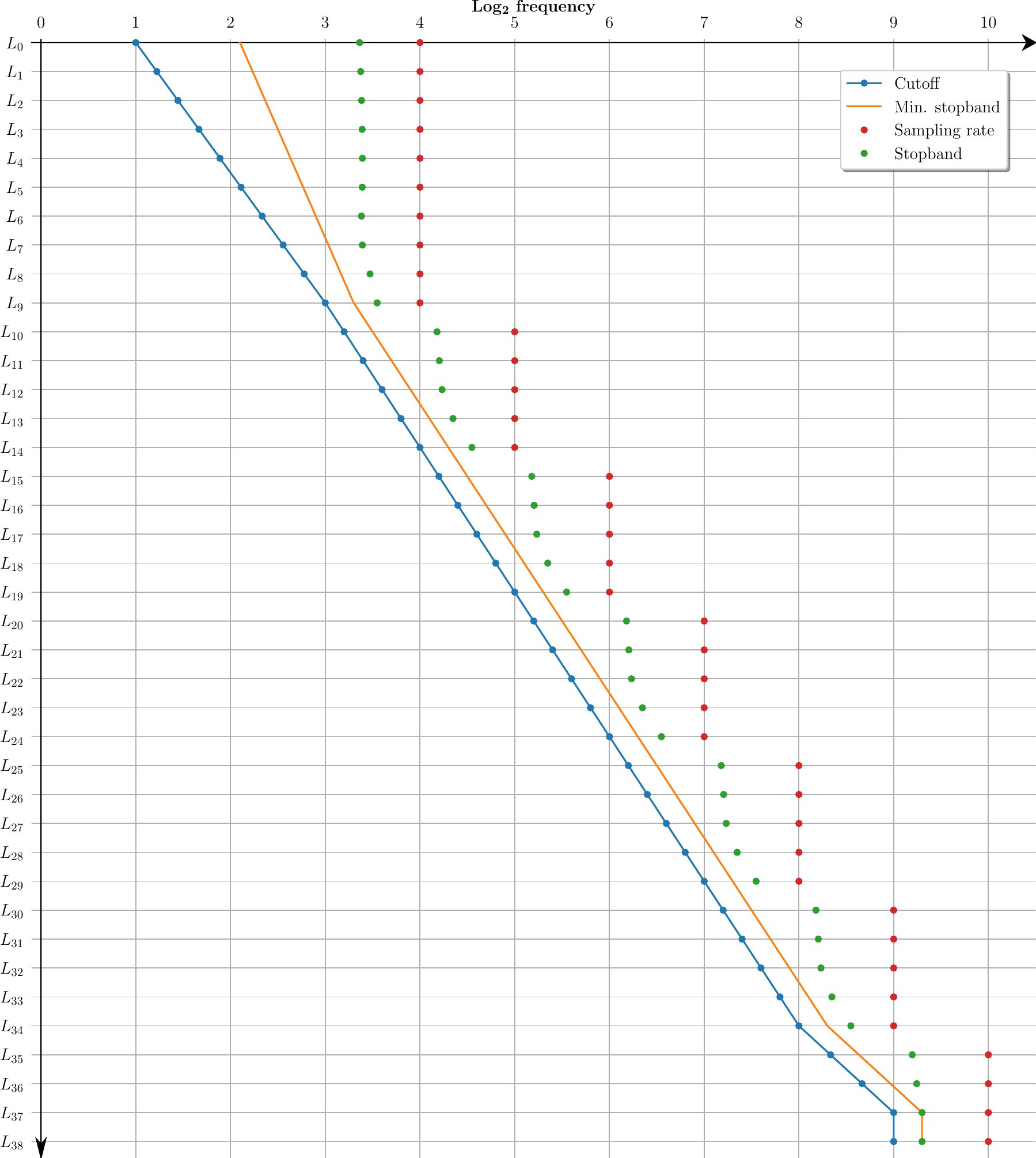}
  \caption{\textbf{Flexible Layer Specification of Stylegan-XL}. StyleGAN-XL consists of $39$ layers at resolution $1024^2$. Cutoff (blue) and minimum acceptable stopband frequency (orange) obey geometric progression over the layers; sampling rate (red) and actual stopband (green) are computed according to our design constraints.}
  \label{fig:layerspecs}
\end{figure*}
}

\newcommand{\aquamen}{
\begin{figure}
  \includegraphics[width=1.0\linewidth]{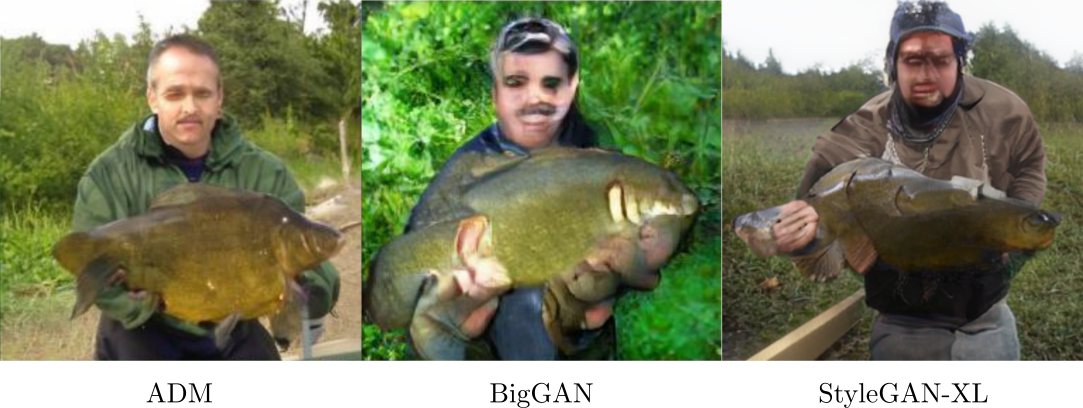}
  \caption{\textbf{Imagenet Classes Containing Humans}.   
  Samples for BigGAN and ADM are taken from~\cite{Dhariwal2021NEURIPS}.
  }
  \label{fig:aquamen}
\end{figure}
}

\newcommand{\interpsupp}{
\begin{figure*}[!p]
  \includegraphics[width=\textwidth]{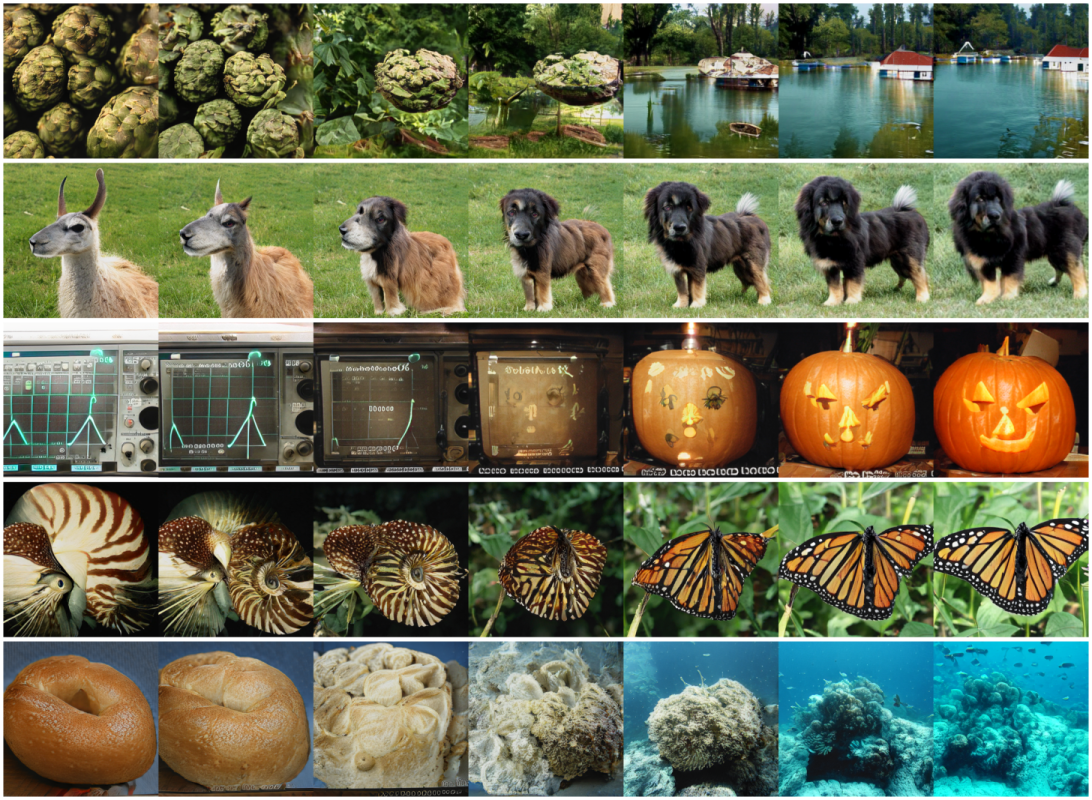}
  \caption{
  \textbf{Interpolations.} StyleGAN-XL generates smooth interpolations between samples of different classes.
  }
  \label{fig:interpsupp}
\end{figure*}
}

\newcommand{\editingsupp}{
\begin{figure*}[!p]
  \includegraphics[width=\textwidth]{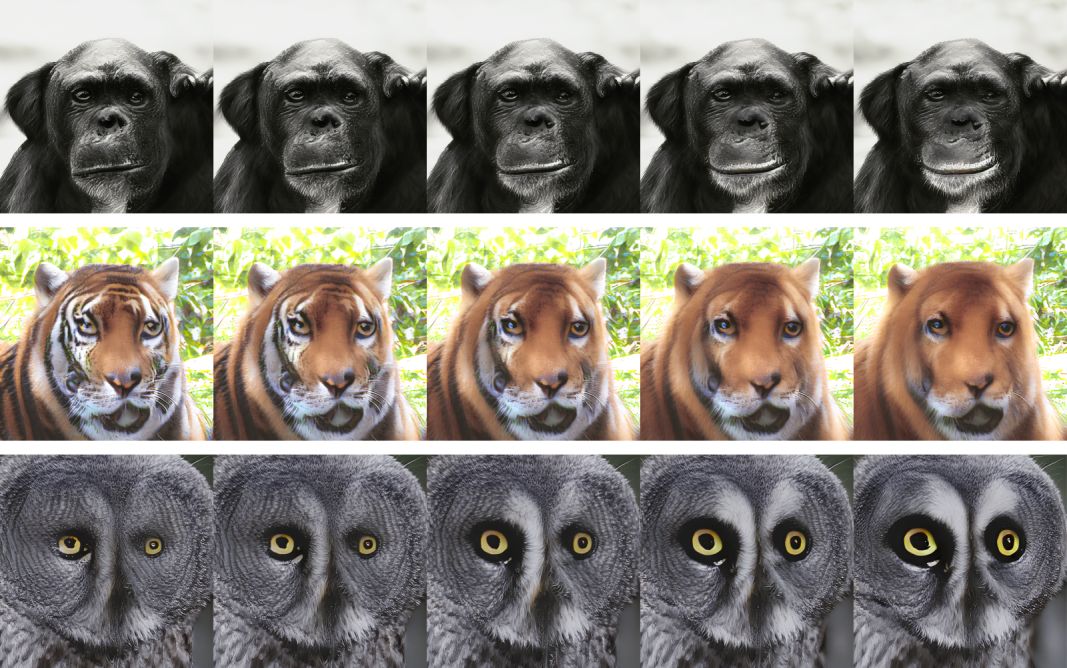}
  \caption{\textbf{Image Manipulation via Language}. 
 Given a random sample, we manipulate the image by by following semantic directions in latent space found by StyleMC~\cite{Kocasari2021WACV}. The latent space directions from top to bottom are: "smile", "no stripes", and "big eyes".
  }
  \label{fig:editingsupp}
\end{figure*}
}

\newcommand{\ffhqsamples}{
\begin{figure*}[!p]
  \includegraphics[height=0.95\textheight]{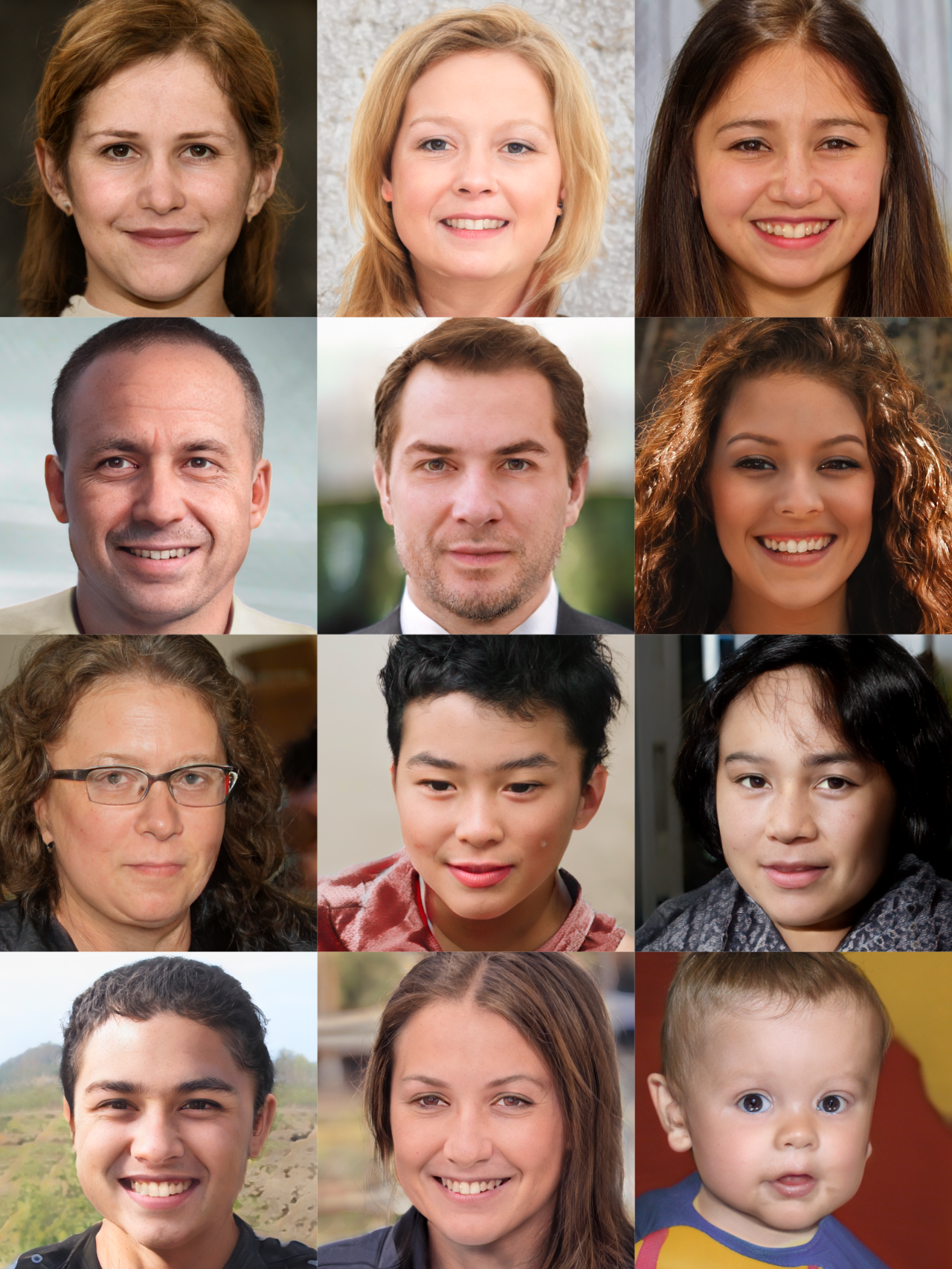}
  \caption{\textbf{Samples on FFHQ $1024^2$}. 
  }
  \label{fig:ffhqsamples}
\end{figure*}
}

\newcommand{\pokemonsamples}{
\begin{figure*}[!p]
  \includegraphics[height=0.95\textheight]{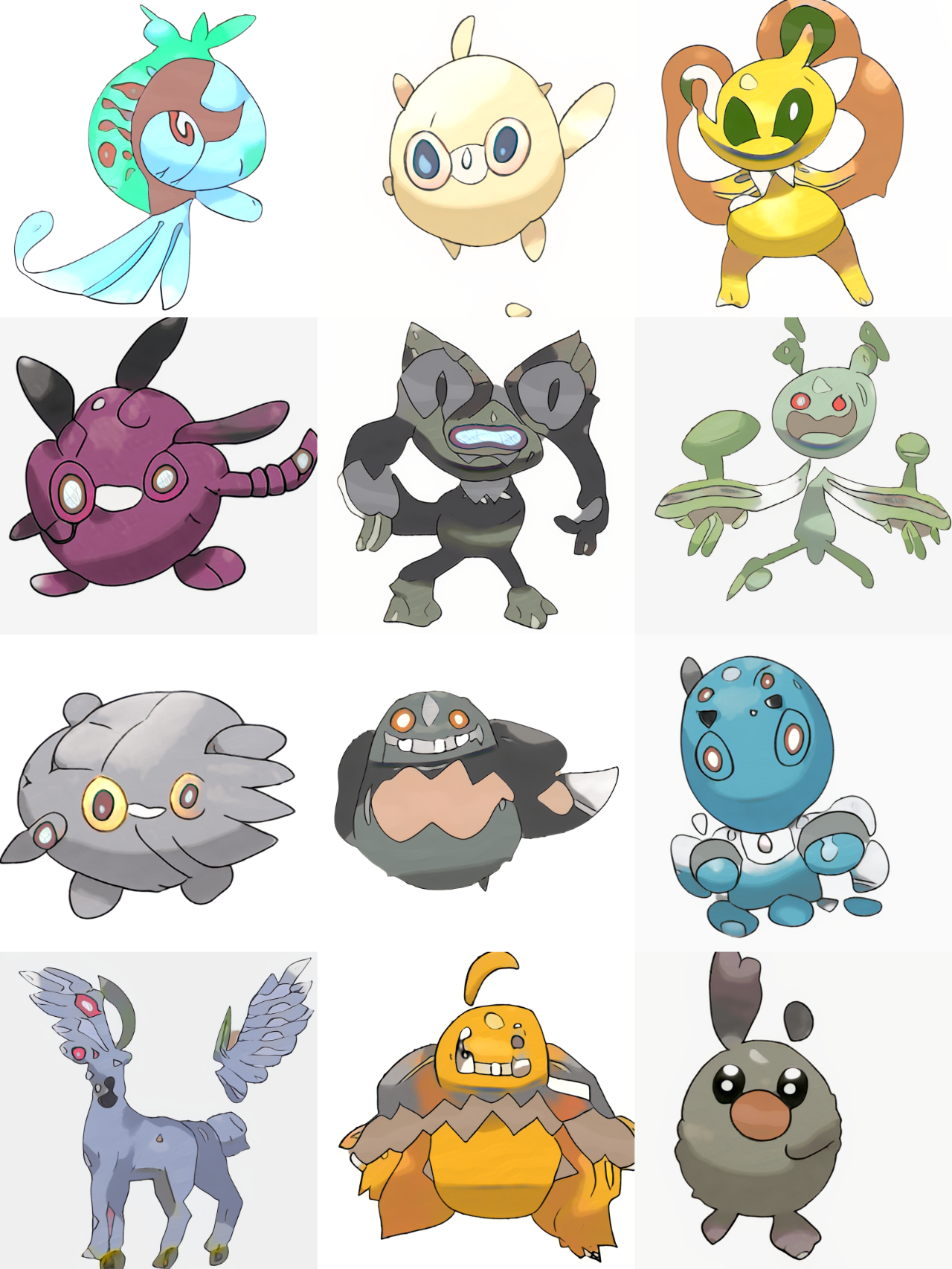}
  \caption{\textbf{Samples on Pokemon $1024^2$}. 
  }
  \label{fig:pokemonsamples}
\end{figure*}
}

\newcommand{\samplesa}{
\begin{figure*}[!p]
  \includegraphics[width=\textwidth]{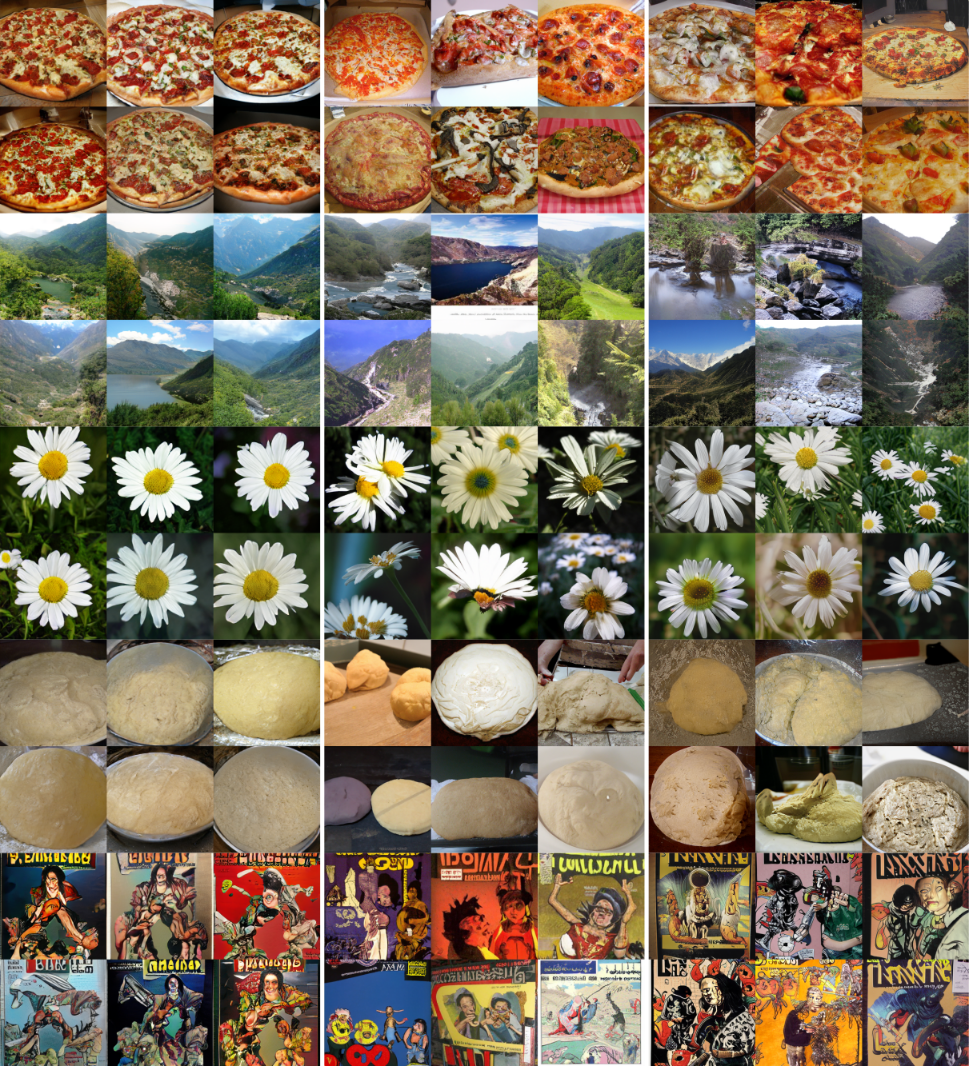}
  \caption{\textbf{Qualitiative Comparison on ImageNet $256^2$.}. 
  We compare BigGAN (left column), ADM (middle column), and StyleGAN-XL (right column). Classes from top to bottom: pizza, valley, daisy, dough, comic book.
  }
  \label{fig:samplesa}
\end{figure*}
}

\newcommand{\samplesb}{
\begin{figure*}[!p]
  \includegraphics[width=\textwidth]{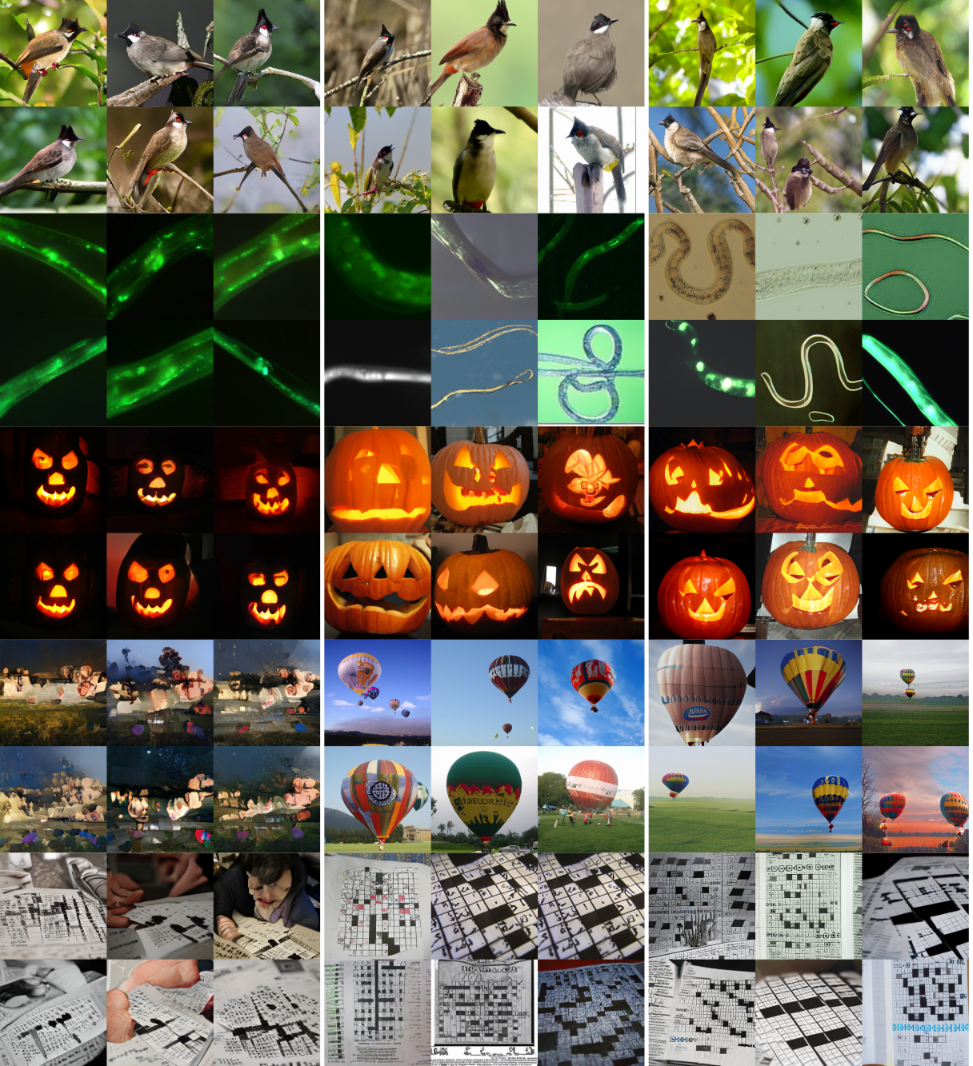}
  \caption{\textbf{Qualitiative Comparison on ImageNet $256^2$.}. 
  We compare BigGAN (left column), ADM (middle column), and StyleGAN-XL (right column). Classes from top to bottom: bulbul, nematode, jack-o'-lantern, balloon, crossword puzzle.
  }
  \label{fig:samplesb}
\end{figure*}
}

\newcommand{\samplesc}{
\begin{figure*}[!t]
  \includegraphics[width=\textwidth]{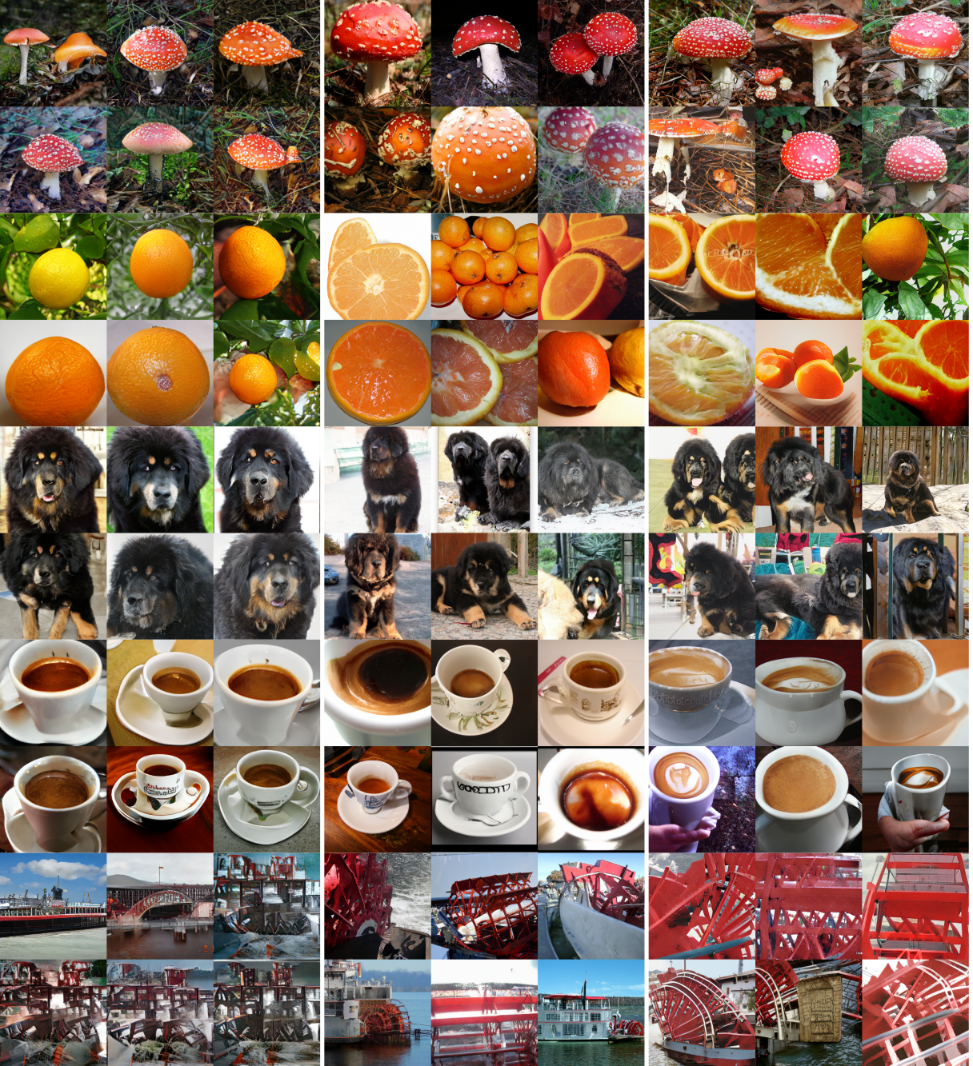}
  \caption{\textbf{Qualitiative Comparison on ImageNet $256^2$.}. 
  We compare BigGAN (left column), ADM (middle column), and StyleGAN-XL (right column). Classes from top to bottom: agaric, orange, Tibetian mastiff, espresso, paddlewheel.
  }
  \label{fig:samplesc}
\end{figure*}
}
\providecommand{\impath}[1]{}
\providecommand{\impatha}[1]{}
\providecommand{\impathb}[1]{}
\providecommand{\impathc}[1]{}
\providecommand{\impathd}[1]{}
\providecommand{\impathe}[1]{}

\newcommand{\sotapr}{
\begin{figure*}[t]
\captionof{table}{\textbf{Image Synthesis on ImageNet}. Empty cells indicate that the model was not available and the respective metric not evaluated in the original work.
}
\small
\resizebox{\textwidth}{!}{
\begin{tabular}{@{}lrrrrrrclrrrrrr@{}}
\toprule
\textbf{Model} & \textbf{FID $\downarrow$} & \textbf{sFID $\downarrow$} & \textbf{rFID $\downarrow$} & \textbf{IS $\uparrow$} & \textbf{Pr $\uparrow$} & \textbf{Rec $\uparrow$} & \textbf{} & \textbf{Model} & \textbf{FID $\downarrow$} & \textbf{sFID $\downarrow$} & \textbf{rFID $\downarrow$} & \textbf{IS $\uparrow$} & \textbf{Pr $\uparrow$} & \textbf{Rec $\uparrow$} \\
\textbf{\rule{0pt}{4ex}Resolution $128^2$} &  &  &  &  &  &  &  & \textbf{Resolution $256^2$} &  &  &  &  &  &  \\ \midrule
BigGAN & 6.02 & 7.18 & 6.09 & 145.83 & \textbf{0.86} & 0.35 & \multicolumn{1}{l}{} & StyleGAN2 & 49.20 &  &  &  &  &  \\
CDM & 3.52 & 128.80 &  & 128.80 &  &  &  & BigGAN & 6.95 & 7.36 & 75.24 & 202.65 & \textbf{0.87} & 0.28 \\
ADM & 5.91 & 5.09 & 13.29 & 93.31 & 0.70 & \textbf{0.65} &  & CDM & 4.88 & 158.70 &  & 158.70 &  &  \\
ADM-G & 2.97 & 5.09 & 3.80 & 141.37 & 0.78 & 0.59 &  & ADM & 10.94 & 6.02 & 125.78 & 100.98 & 0.69 & \textbf{0.63} \\
StyleGAN-XL & \textbf{1.81} & \textbf{3.82} & \textbf{1.82} & \textbf{200.55} & 0.77 & 0.55 &  & ADM-G-U & 3.94 & 6.14 & 11.86 & 215.84 & 0.83 & 0.53 \\
 &  &  &  &  &  &  &  & StyleGAN-XL & \textbf{2.30} & \textbf{4.02} & \textbf{7.06} & \textbf{265.12} & 0.78 & 0.53 \\
\textbf{\rule{0pt}{4ex}Resolution $512^2$} & \textbf{} & \textbf{} & \textbf{} & \textbf{} & \textbf{} & \textbf{} & \textbf{} & \textbf{Resolution $1024^2$} & \textbf{} & \textbf{} & \textbf{} & \textbf{} & \textbf{} & \textbf{} \\ \midrule
BigGAN & 8.43 & 8.13 & 312.00 & 177.90 & \textbf{0.88} & 0.29 &  & StyleGAN-XL & \textbf{2.52} & \textbf{4.12} & \textbf{413.12} & \textbf{260.14} & \textbf{0.76} & \textbf{0.51} \\
ADM & 23.24 & 10.19 & 561.32 & 58.06 & 0.73 & \textbf{0.60} &  &  &  &  &  &  &  &  \\
ADM-G-U & 3.85 & 5.86 & 210.83 & 221.72 & 0.84 & 0.53 &  &  &  &  &  &  &  &  \\
StyleGAN-XL & \textbf{2.41} & \textbf{4.06} & \textbf{51.54} & \textbf{267.75} & 0.77 & 0.52 &  &  &  &  &  &  &  &  \\ \bottomrule
\end{tabular}
}  
\label{tab:sotapr}  
\end{figure*}
}

\newcommand{\ablation}{
\begin{figure*}
\captionof{table}{\textbf{Ablation Study on ImageNet $128^2$}. Left: Results for different configurations after training for $15$ V100-days. Right: Comparing combinations of different feature networks $\bF$. Beginning from the base configuration using an EfficientNet-lite0 (EffNet), we add a second $\bF$ with varying architecture type and pretraining objective (\textit{Class}: Classification, \textit{Self}: MoCo-v2~\cite{Chen2020ARXIV}).
}
\begin{minipage}[b][][b]{.45\textwidth}
\centering
\begin{tabular}{@{}clccc@{}}
\toprule
\multicolumn{2}{l}{\textbf{Configuration}} & \textbf{FID} $\downarrow$ & \textbf{IS} $\uparrow$\\ \midrule
\textbf{\;\;A} & StyleGAN3                      & 53.57 & 15.30  \\
\textbf{\;\;B} & + Projected GAN \& small $\bz$ & 22.98 & 57.62  \\
\textbf{\;\;C} & + Pretrained embeddings        & 20.91 & 35.79  \\
\textbf{\;\;D} & + Progressive growing          & 19.51 & 35.74  \\
\textbf{\;\;E} & + ViT \& CNN as $\bF_{1,2}$    & 12.43 & 56.72   \\
\textbf{\;\;F} & + CLF guidance (StyleGAN-XL)   & \textbf{12.24}  & \textbf{86.21} \\ \bottomrule
\end{tabular}
\end{minipage} \quad
\begin{minipage}[b][][b]{.45\textwidth}
\centering
\renewcommand{\arraystretch}{1.17}
\begin{tabular}{@{}cccccccc@{}}
\toprule
\multicolumn{2}{c}{Model} & \multicolumn{2}{c}{Type} & \multicolumn{2}{c}{Objective} & \textbf{FID $\downarrow$} & \textbf{IS $\uparrow$} \\ 
$\bF_1$ & $\bF_2$ & $\bF_1$ & $\bF_2$ & $\bF_1$ & $\bF_2$ &  &  \\ \midrule
EffNet &          & CNN &     & Class   &        & 19.51 & 35.74 \\
EffNet & ResNet50 & CNN & CNN & Class   & Class    & 16.16 & 49.13 \\
EffNet & ResNet50 & CNN & CNN & Class   & Self  & 18.53 & 38.26 \\
EffNet & DeiT-M   & CNN & ViT & Class   & Class    & \textbf{12.43} & \textbf{56.72} \\ \bottomrule
\end{tabular}
\end{minipage}
\label{tab:ablation}
\end{figure*}
}

\newcommand{\invresults}{
\begin{figure}[]
\captionof{table}{\textbf{Inversion Results.} The metrics are computed between the inversions obtained by the model and the reconstruction targets.}
\begin{tabular}{@{}llccc@{}}
\toprule
\textbf{Model} & \textbf{MSE} $\downarrow$ & \textbf{PSNR} $\uparrow$ & \textbf{SSIM} $\uparrow$ & \textbf{FID} $\downarrow$\\ \midrule
BigGAN & 0.10 & 10.85 & 0.26 & 47.48 \\
StyleGAN-XL & \textbf{0.06} & \textbf{13.45} & \textbf{0.33} & \textbf{21.73} \\ \bottomrule
\end{tabular}
\label{tab:invresults}
\end{figure}
}

\newcommand{\imagenetlowres}{
\begin{figure}[]
\captionof{table}{\textbf{Results on ImageNet at Lower Resolutions.}.}
\begin{tabular}{lrrr}
\hline
\rule{0pt}{3ex}\textbf{Model} & \multicolumn{3}{c}{\textbf{FID $\downarrow$}} \\
\rule{0pt}{3ex} & \textbf{Res. $16^2$} & \textbf{Res. $32^2$} & \textbf{Res. $64^2$} \\ \hline
\rule{0pt}{3ex}StyleGAN-XL & 0.73 & 1.10 & 1.51 \\ \hline
\end{tabular}
\label{tab:imagenetlowres}
\end{figure}
}

\newcommand{\extraresults}{
\begin{figure}[]
\captionof{table}{\textbf{Results on Unimodal Datasets.}.}
\begin{tabular}{@{}lllll@{}}
\toprule
\textbf{Model} & \textbf{FID} & \textbf{} & \textbf{Model} & \textbf{FID} \\
\textbf{\rule{0pt}{4ex} FFHQ $1024^2$} &  &  & \textbf{Pok\'emon $1024^2$} &  \\ \midrule
StyleGAN2 & 2.70 &  & FastGAN & 56.46 \\
StyleGAN3 & 2.79 &  & Projected GAN & 33.96 \\
StyleGAN-XL & \textbf{2.02} &  & StyleGAN-XL & \textbf{25.47} \\ \bottomrule
\end{tabular}
\label{tab:extraresults}
\end{figure}
}

\newcommand{\speedcomparison}{
\begin{figure}[]
\captionof{table}{\textbf{Inference speed comparison.}. We measure the time required for a forward pass with batch size $1$ in V100-seconds. ADM uses classifier guidance.}
\begin{tabular}{lrrr}
\hline
\textbf{\rule{0pt}{3ex}\textbf{Model}} & \multicolumn{3}{c}{\textbf{Inference Time $\downarrow$}} \\
\rule{0pt}{3ex} & \textbf{\textbf{Res. $128^2$}} & \textbf{\textbf{Res. $256^2$}} & \textbf{\textbf{Res. $512^2$}} \\ \hline
\rule{0pt}{3ex}ADM & 27.07 & 40.26 & 91.54 \\
StyleGAN-XL & 0.05 & 0.07 & 0.10 \\ \hline
\end{tabular}
\label{tab:speedcomparison}
\end{figure}
}
\begin{document}
\title{StyleGAN-XL: Scaling StyleGAN to Large Diverse Datasets}

\author{Axel Sauer}
\email{a.sauer@uni-tuebingen.de}

\author{Katja Schwarz}
\email{katja.schwarz@uni-tuebingen.de}

\author{Andreas Geiger}
\affiliation{%
  \institution{\\University of T{\"u}bingen and Max Planck Institute for Intelligent Systems, T{\"u}bingen}
  \country{Germany}
}
\email{a.geiger@uni-tuebingen.de}

\renewcommand\shortauthors{Sauer et al.}

\begin{abstract}
Computer graphics has experienced a recent surge of data-centric approaches for photorealistic and controllable content creation. StyleGAN  in particular sets new standards for generative modeling regarding image quality and controllability. However, StyleGAN's performance severely degrades on large unstructured datasets such as ImageNet. StyleGAN was designed for controllability; hence, prior works suspect its restrictive design to be unsuitable for diverse datasets. In contrast, we find the main limiting factor to be the current training strategy. Following the recently introduced Projected GAN paradigm, we leverage powerful neural network priors and a progressive growing strategy to successfully train the latest StyleGAN3 generator on ImageNet. Our final model, StyleGAN-XL, sets a new state-of-the-art on large-scale image synthesis and is the first to generate images at a resolution of $1024^2$ at such a dataset scale. We demonstrate that this model can invert and edit images beyond the narrow domain of portraits or specific object~classes.
Code, models, and supplementary videos can be found at \texttt{\url{https://sites.google.com/view/stylegan-xl/}}.
\end{abstract} 



\teaser

\maketitle

\section{Introduction}
\label{sec:intro}

Computer graphics has long been concerned with generating photorealistic images at high resolution that allow for direct control over semantic attributes. Until recently, the primary paradigm was to create carefully designed 3D models which are then rendered using realistic camera and illumination models. A parallel line of research approaches the problem from a data-centric perspective. 
In particular, probabilistic generative models ~\cite{Goodfellow2014NEURIPS,Oord2017NEURIPS,Song2021ICLR} have shifted the paradigm from designing assets to designing training procedures and datasets. Style-based GANs (StyleGANs) are a specific instance of these models, and they exhibit many desirable properties. They achieve high image fidelity~\cite{Karras2019CVPR, Karras2020CVPR}, fine-grained semantic control~\cite{Haerkoenen2020NEURIPS, WU2021CVPRa,Ling2021ARXIV}, and recently alias-free generation enabling realistic animation~\cite{Karras2021NEURIPS}. Moreover, they reach impressive photorealism on carefully curated datasets, especially of human faces. However, when trained on large and unstructured datasets like ImageNet~\cite{Deng2009CVPR}, StyleGANs do not achieve satisfactory results yet. One other problem plaguing data-centric methods, in general, is that they become prohibitively more expensive when scaling to higher resolutions as bigger models are required.

Initially, StyleGAN~\cite{Karras2019CVPR} was proposed to explicitly disentangle factors of variations, allowing for better control and interpolation quality. 
However, its architecture is more restrictive than a standard generator network~\cite{Radford2016ICLR, Karras2018ICLR} which seems to come at a price when training on complex and diverse datasets such as ImageNet. Previous attempts at scaling StyleGAN and StyleGAN2 to ImageNet led to sub-par results~\cite{Gwern2020MISC, Grigoryev2022ICLR},  giving reason to believe it might be fundamentally limited for highly diverse datasets~\cite{Gwern2020MISC}.

BigGAN~\cite{Brock2019ICLR} is the state-of-the-art GAN model for image synthesis on ImageNet. The main factors for BigGANs success are larger batch and model sizes.
However, BigGAN has not reached a similar standing as StyleGAN as its performance varies significantly between training runs~\cite{Karras2020NeurIPS} and as it does not employ an intermediate latent space which is essential for GAN-based image editing~\cite{Abdal2021TOG, Patashnik2021ICCV, Collins2020CVPR, WU2021CVPRa}. Recently, BigGAN has been superseded in performance by diffusion models~\cite{Dhariwal2021NEURIPS}. Diffusion models achieve more diverse image synthesis than GANs but are significantly slower during inference and prior work on GAN-based editing is not directly applicable. Following these arguments, successfully training StyleGAN on ImageNet has several advantages over existing methods.

The previously failed attempts at scaling StyleGAN raise the question of whether architectural constraints fundamentally limit style-based generators or if the missing piece is the right training strategy.
Recent work by \cite{Sauer2021NEURIPS} introduced \textit{Projected GANs} which project generated and real samples into a fixed, pretrained feature space. Rephrasing the GAN setup this way leads to significant improvements in training stability, training time, and data efficiency.
Leveraging the benefits of Projected GAN training might enable scaling StyleGAN to ImageNet.
However, as observed by~\cite{Sauer2021NEURIPS}, the advantages of Projected GANs only partially extend to StyleGAN on the unimodal datasets they investigated.
We study this issue and propose architectural changes to address it.
We then design a progressive growing strategy tailored to the latest StyleGAN3. 
These changes in conjunction with Projected GAN already allow surpassing prior attempts of training StyleGAN on ImageNet. 
To further improve results, we analyze the pretrained feature network used for Projected GANs and find that the two standard neural architectures for computer vision, CNNs and ViTs~\cite{Dosovitskiy2021ICLR}, significantly improve performance when used jointly. Lastly, we leverage \textit{classifier guidance}, a technique originally introduced for diffusion models to inject additional class-information~\cite{Dhariwal2021NEURIPS}.

Our contributions culminate in a new state-of-the-art on large-scale image synthesis, pushing the performance beyond existing GAN and diffusion models. 
We showcase inversion and editing for ImageNet classes and find that Pivotal Tuning Inversion (PTI)~\cite{Roich2021ARXIV}, a powerful new inversion paradigm, combines well with our model and even embeds out-of-domain images smoothly into our learned latent space. Our efficient training strategy allows us to triple the parameters of the standard StyleGAN3 while reaching prior state-of-the-art performance of diffusion models~\cite{Dhariwal2021NEURIPS} in a fraction of their training time. It further enables us to be the first to demonstrate image synthesis on ImageNet-scale at a resolution of $1024^2$ pixels. We will open-source our code and models upon publication. 

\section{Background}
\label{sec:background}
We first introduce the main building blocks of our system:
the StyleGAN3 generator~\cite{Karras2021NEURIPS} and Projected GAN's~\cite{Sauer2021NEURIPS} feature projectors and multi-scale discriminators.

\boldparagraph{StyleGAN.}
This section describes style-based generators in general with a focus on the latest StyleGAN3~\cite{Karras2021NEURIPS}. A StyleGAN generator consists of a mapping network $\bG_m$ and a synthesis network $\bG_s$. First, $\bG_m$ maps a normally distributed latent code $\bz$ to a style code $\bw$. This style code $\bw$ is then used for modulating the convolution kernels of $\bG_s$ to control the synthesis process.
The synthesis network $\bG_s$ of StyleGAN3 starts from a spatial map defined by Fourier features~\cite{Tancik2020NEURIPS, Xu2021CVPR}. This input then passes through $N$ layers of convolutions, non-linearities, and upsampling to generate an image. 
Each non-linearity is wrapped by an upsampling and downsampling operation to prevent aliasing. 
The low-pass filters used for these operations are carefully designed to
balance image quality, antialiasing, and training speed.
Concretely, their cutoff and stopband frequencies grow geometrically with network depth, the transition band half-widths are as wide as possible within the limits of the layer sampling rate, and only the last two layers are critically sampled, i.e., the filter cutoff equals the bandlimit. The number of layers $N$ is $14$, independent of the final output resolution.

Style mixing and path length regularization are methods for regularizing style-based generators. In style mixing, an image is generated by feeding sampled style codes $\bw$ into different layers of $\bG_s$ independently. Path length regularization encourages that a step of fixed size in latent space results in a corresponding fixed change in pixel intensity of the generated image~\cite{Karras2020CVPR}.
This inductive bias leads to a smoother generator mapping and has several advantages including fewer artifacts, more predictable training behavior, and better inversion.

Progressive growing was introduced by~\cite{Karras2018ICLR} for stable training at high resolutions but~\cite{Karras2020CVPR} found that it can impair shift-equivariance.~\cite{Karras2021NEURIPS} observe that texture sticking artifacts are caused by a lack of equivariance and carefully design StyleGAN3 to prevent texture sticking. Hence, in this paper, as we build on StyleGAN3, we can revisit the idea of progressive growing to improve convergence speed and synthesis quality.

\boldparagraph{Projected GAN.}
The original adversarial game between a generator $\bG$ and a discriminator $\bD$ can be extended by a set of feature projectors $\{\bP_l\}$~\cite{Sauer2021NEURIPS}. The projectors map real images $\bx$ and images generated by $\bG$ to the discriminator's input space. 
The Projected GAN objective is formulated as
\begin{equation}
\begin{aligned}
    \min_\bG \max_{\{\bD_l\}} &\sum_{l \in \cL} \Big (
    \nE_{\bx} [\log \bD_l(\bP_l(\bx))]\\ 
    &\quad \; + \nE_{\bz}[ \log( 1- \bD_l(\bP_l(\bG(\bz))))] \Big)
\end{aligned}
\end{equation}
\label{eq:GANobjective2}

where $\{\bD_l\}$ is a set of independent discriminators operating on different feature projections.
The projectors consist of a pretrained feature network $\bF$, cross-channel mixing (CCM) and cross-scale mixing (CSM) layers. 
The purpose of CCM and CSM is to prohibit the discriminators from focusing on only a subset of its input feature space which would result in mode collapse. Both modules employ differentiable random projections that are not optimized during GAN training. CCM mixes features across channels via random 1x1 convolutions, CSM mixes features across scales via residual random 3x3 convolution blocks and bilinear upsampling. The output of CSM is a feature pyramid consisting of four feature maps at different resolutions. Four discriminators operate independently on these feature maps. Each discriminator uses a simple convolutional architecture and spectral normalization~\cite{Miyato2018ICLR}. The depth of the discriminator varies depending on its input resolution, i.e., a spatially larger feature map corresponds to a deeper discriminator. Other than spectral normalization, Projected GANs do not use additional regularization such as gradient penalties~\cite{Mescheder2018ICML}. Lastly, ~\cite{Sauer2021NEURIPS} apply differentiable data-augmentation \cite{Zhao2020NeurIPS} before $\bF$ which improves Projected GAN's performance independent of the dataset size.

~\cite{Sauer2021NEURIPS} evaluate several combinations of $\bF$ and $\bG$ and find an EfficientNet-Lite0~\cite{Tan2019ICML} and a FastGAN generator~\cite{Liu2021ICLR} to work especially well. When using a StyleGAN generator, they observe that the discriminators can quickly overpower the generator for suboptimal learning rates. The authors suspect that the generator might adapt too slowly due to its design which modulates feature maps with styles learned by a mapping network.

\section{Scaling StyleGAN to ImageNet}
\label{sec:method}
As mentioned before, StyleGAN has several advantages over existing approaches that work well on ImageNet.
But a na\"{i}ve training strategy does not yield state-of-the-art performance~\cite{Gwern2020MISC, Grigoryev2022ICLR}.
Our experiments confirm that even the latest StyleGAN3 does not scale well, see \figref{fig:teaser}. Particularly at high resolutions, the training becomes unstable.
Therefore, our goal is to train a StyleGAN3 generator on ImageNet successfully. Success is defined in terms of sample quality primarily measured by inception score (IS)~\cite{Salimans2016NEURIPS} and diversity measured by Fr\'echet Inception Distance (FID)~\cite{Heusel2017NEURIPS}.
Throughout this section, we gradually introduce changes to the StyleGAN3 baseline (\textbf{Config-A}) and track the improvements in \tabref{tab:ablation}. First, we modify the generator and its regularization losses, adapting the latent space to work well with Projected GAN (\textbf{Config-B}) and for the class-conditional setting (\textbf{Config-C}). We then revisit progressive growing to improve training speed and performance (\textbf{Config-D}). Next, we investigate the feature networks used for Projected GAN training to find a well-suited configuration (\textbf{Config-E}). Lastly, we propose classifier guidance for GANs to provide class information via a pretrained classifier (\textbf{Config-F}). Our contributions enable us to train a significantly larger model than previously possible while requiring less computation than prior art. Our model is three times larger in terms of depth and parameter count than a standard StyleGAN3. However, to match the prior state-of-the-art performance of ADM~\cite{Dhariwal2021NEURIPS} at a resolution of $512^2$ pixels, training the models on a single NVIDIA Tesla V100 takes $400$ days compared to the previously required $1914$ V100-days. We refer to our model as \textbf{StyleGAN-XL} (\figref{fig:system}).
\system
\ablation

\subsection{Adapting Regularization and Architectures}
Training on a diverse and class-conditional dataset makes it necessary to introduce several adjustments to the standard StyleGAN configuration. We construct our generator architecture using layers of StyleGAN3-T, the translational-equivariant configuration of StyleGAN3. In initial experiments, we found the rotational-equivariant StyleGAN3-R to generate overly symmetric images on more complex datasets, resulting in kaleidoscope-like patterns. 
 
\boldparagraph{Regularization.}
In GAN training, it is common to use regularization for both, the generator and the discriminator. Regularization improves results on uni-modal datasets like FFHQ~\cite{Karras2019CVPR} or LSUN~\cite{Yu2015ARXIV}, whereas it can be detrimental on multi-modal datasets~\cite{Brock2019ICLR, Gwern2020MISC}. Therefore, we aim to avoid regularization when possible.~\cite{Karras2021NEURIPS} find style mixing to be unnecessary for the latest StyleGAN3; hence, we also disable it. Path length regularization can lead to poor results on complex datasets~\cite{Gwern2020MISC} and is, per default, disabled for StyleGAN3~\cite{Karras2021NEURIPS}. However, path length regularization is attractive as it enables high-quality inversion~\cite{Karras2020CVPR}. 
We also observe unstable behavior and divergence when using path length regularization in practice. We found that this problem can be circumvented by only applying regularization after the model has been sufficiently trained, i.e., after 200k images. For the discriminator, following~\cite{Sauer2021NEURIPS}, we use spectral normalization without gradient penalties. In addition, we blur all images with a Gaussian filter with $\sigma=2$ pixels for the first $200k$ images. Discriminator blurring has been introduced in ~\cite{Karras2021NEURIPS} for StyleGAN3-R. It prevents the discriminator from focusing on high frequencies early on, which we found beneficial across all settings we investigated. 
 
\boldparagraph{Low-Dimensional Latent Space.}
As observed in~\cite{Sauer2021NEURIPS}, Projected GANs work better with FastGAN~\cite{Liu2021ICLR} than with StyleGAN. One main difference between these generators is their latent space, StyleGAN's latent space is comparatively high dimensional 
(FastGAN: $\mathbb{R}^{100}$, BigGAN: $\mathbb{R}^{128}$, StyleGAN: $\mathbb{R}^{512}$). 
Recent findings indicate that the \textit{intrinsic dimension}  of natural image datasets is relatively low~\cite{Pope2021ICLR}, ImageNet's dimension estimate is around $40$. Accordingly, a latent code of size $512$ is highly redundant, making the mapping network's task harder at the beginning of training. Consequently, the generator is slow to adapt and cannot benefit from Projected GAN's speed up. We therefore reduce StyleGAN's latent code $\bz$ to $64$ and now observe stable training in combination with Projected GAN, resulting in lower FID than the baseline (\textbf{Config-B}). We keep the original dimension of the \textit{style code} $\bw \in \mathbb{R}^{512}$ to not restrict the model capacity of the mapping network $\bG_m$.
 
\boldparagraph{Pretrained Class Embeddings.} 
Conditioning the model on class information is essential to control the sample class and improve overall performance. A class-conditional variant of StyleGAN was first proposed in~\cite{Karras2020NeurIPS} for CIFAR10~\cite{Krizhevsky2009CITESEER} where a one-hot encoded label is embedded into a 512-dimensional vector and concatenated with $\bz$. For the discriminator, class information is projected onto the last discriminator layer~\cite{Miyato2018ICLRb}. We observe that \textbf{Config-B} tends to generate similar samples per class resulting in high IS.
To quantify mode coverage, we leverage the recall metric~\cite{Kynknniemi2019NEURIPS} and find that \textbf{Config-B} achieves a low recall of~$0.004$. We hypothesize that the class embeddings collapse when training with Projected GAN. Therefore, to prevent this collapse, we aim to ease optimization of the embeddings via pretraining. We extract and spatially pool the lowest resolution features of an Efficientnet-lite0~\cite{Tan2019ICML} and calculate the mean per ImageNet class. 
The network has a low channel count to keep the embedding dimension small, following the arguments of the previous section. The embedding passes through a linear projection to match the size of $\bz$ to avoid an imbalance. Both $\bG_m$ and $\bD_i$ are conditioned on the embedding.
During GAN training, the embedding and the linear projection are optimized to allow specialization. Using this configuration, we observe that the model generates diverse samples per class, and recall increases to $0.15$ (\textbf{Config-C}). Note that for all configurations in this ablation, we restrict the training time to $15\;V\text{-}100\;days$. Hence, the absolute recall is markedly lower compared to the fully trained models. Conditioning a GAN on pretrained features was also recently investigated by~\cite{Casanova2021NEURIPS}. In contrast to our approach,~\cite{Casanova2021NEURIPS} condition on specific \textit{instances}, instead of learning a general class embedding.

\subsection{Reintroducing Progressive Growing}
Progressively growing the output resolution of a GAN was introduced by~\cite{Karras2018ICLR} for fast and more stable training. The original formulation adds layers during training to both $\bG$ and $\bD$ and gradually fades in their contribution. However, in a later work, it was discarded~\cite{Karras2020CVPR} as it can contribute to texture sticking artifacts. Recent work by~\cite{Karras2021NEURIPS} finds that the primary cause of these artifacts is aliasing, so they redesign each layer of StyleGAN to prevent it. This motivates us to reconsider progressive growing with a carefully crafted strategy that aims to suppress aliasing as best as possible. Training first on very low resolutions, as small as $16^2$ pixels, enables us to break down the daunting task of training on high-resolution ImageNet into smaller subtasks. This idea is in line with the latest work on diffusion models~\cite{Nichol2021ICML, Saharia2021ARXIV, Dhariwal2021NEURIPS, Ho2022JMLR}. They observe considerable improvements in FID on ImageNet when using a two-stage model, i.e., stacking an independent low-resolution model and an upsampling model to generate the final image.

Commonly, GANs follow a rigid sampling rate progression, i.e., at each resolution, there is a fixed amount of layers followed by an upsampling operation using fixed filter parameters. StyleGAN3 does not follow such a progression. Instead, the layer count is set to $14$, independent of the output resolution, and the filter parameters of up- and downsampling operations are carefully designed for antialiasing under the given configuration. The last two layers are critically sampled to generate high-frequency details. When adding layers for the subsequent highest resolution, discarding the previously critically sampled layers is crucial as they would introduce aliasing when used as intermediate layers~\cite{Karras2020CVPR, Karras2021NEURIPS}. Furthermore, we adjust the filter parameters of the added layers to adhere to the flexible layer specification of~\cite{Karras2021NEURIPS}; we refer to the supplementary for details. In contrast to~\cite{Karras2018ICLR} we do not add layers to the discriminator. Instead, to fully utilize the pretrained feature network $\bF$, we upsample both data and synthesized images to $\bF$'s training resolution ($224^2$ pixels) when training on smaller images.

We start progressive growing at a resolution of $16^2$ using $11$ layers. Every time the resolution increases, we cut off $2$ layers and add $7$ new ones. Empirically, fewer layers result in worse performance; adding more leads to increased overhead and diminishing returns. For the final stage at $1024^2$, we add only $5$ layers as the last two are not discarded. This amounts to $39$ layers at the maximum resolution of $1024^2$. Instead of a fixed growing schedule, each stage is trained until FID stops decreasing. We find it beneficial to use a large batch size of $2048$ on lower resolution ($16^2$ and $32^2$), similar to~\cite{Brock2019ICLR}. On higher resolutions, smaller batch sizes suffice ($64^2$ to $256^2$: $256$, $512^2$ to $1024^2$: $128$). Once new layers are added, the lower resolution layers remain fixed to prevent mode collapse. 

In our ablation study, FID improves only slightly (\textbf{Config-D}) compared to \textbf{Config-C}. However, the main advantage can be seen at high resolutions, where progressive growing drastically reduces training time. At resolution $512^2$, we reach the prior state-of-the-art (FID$\;=3.85$) after $2$ V100-days.  This reduction is in contrast to other methods such as ADM, where doubling the resolution from $256^2$ to $512^2$ pixels corresponds to increasing training time from $393$ to $1914$ V100-days to find the best performing model\footnote{Note that these settings are not directly comparable as the stem of our model is pretrained, but the values should give a general sense of the order of magnitude.}. As our aim is not to introduce texture sticking artifacts, we measure $EQ\text{-}T$, a metric for determining translation equivariance~\cite{Karras2021NEURIPS}, where higher is better. 
\textbf{Config-C} yields $EQ\text{-}T=55$, while \textbf{Config-D} attains $EQ\text{-}T=48$.
This only slight reduction in equivariance shows that \textbf{Config-D} restricts aliasing almost as well as a configuration without growing. For context, architectures with aliasing yield $EQ\text{-}T\sim 15$.

\subsection{Exploiting Multiple Feature Networks}
An ablation study conducted in~\cite{Sauer2021NEURIPS} finds that most pretrained feature networks $\bF$ perform similarly in terms of FID when used for Projected GAN training regardless of training data, pretraining objective, or network architecture. However, the study does not answer if combining several $\bF$  is advantageous. Starting from the standard configuration, an EfficientNet-lite0, we add a second $\bF$ to inspect the influence of its pretraining objective (classification or self-supervision) and architecture (CNN or Vision Transformer (ViT)~\cite{Dosovitskiy2021ICLR}). The results in \tabref{tab:ablation} show that an additional CNN leads to slightly lower FID. Combining networks with different pretraining objectives does not offer benefits over using two classifier networks. However, combining an EfficientNet with a ViT improves performance significantly. This result corroborates recent results in neural architecture literature, which find that supervised and self-supervised representations are similar~\cite{Grigg2021ARXIV}, whereas ViTs and CNNs learn different representations~\cite{Raghu2021NEURIPS}. Combining both architectures appears to have complementary effects for Projected GANs. We do not see significant improvements when adding more networks; hence, \textbf{Config-E} uses the combination of EfficientNet~\cite{Tan2019ICML} and DeiT-base~\cite{Touvron2021ICML}.

\subsection{Classifier Guidance for GANs}
\cite{Dhariwal2021NEURIPS} introduced classifier guidance to inject class information into diffusion models. Classifier guidance modifies each diffusion step at time step $t$ by adding gradients of a pretrained classifier $\nabla_{\bx_t}\log p_{\phi}(\bc|x_t, t)$. The best results are obtained by applying guidance on class-conditional models and scaling the classifier gradients by a constant $\lambda>1$.  This combination indicates that our model may also profit from classifier guidance, even though it already receives class information via embeddings.

We first pass the generated image $\bx$ through a pretrained classifier CLF to predict the class label $c_i$. We then add a cross-entropy loss $\mathcal{L}_{CE} = -\sum_{i=0}^{C} c_i \log CLF({x}_i) $ as an additional term to the generator loss and scale this term by a constant $\lambda$. For the classifier, we use DeiT-small~\cite{Touvron2021ICML}, which exhibits strong classification performance while not adding much overhead to the training. Similar to ~\cite{Dhariwal2021NEURIPS}, we observe a significant improvement in IS, indicating an increase in sample quality (\textbf{Config-F}).
We find $\lambda=8$ to work well empirically.
Classifier guidance only works well on higher resolutions ($>32^2$); otherwise, it leads to mode collapse. This is in contrast to~\cite{Dhariwal2021NEURIPS} who exclusively guide their low-resolution model. The difference stems from how guidance is applied: we use it for model training, whereas~\cite{Dhariwal2021NEURIPS} guide the sampling process.
\section{Results}
\label{sec:results}
In this section, we first compare StyleGAN-XL to the state-of-the-art approaches for image synthesis on ImageNet. We then evaluate the inversion and editing capabilities of StyleGAN-XL. As described above, we scale our model to a resolution of $1024^2$ pixels, which no prior work has attempted so far on ImageNet. The resolution of most images in ImageNet is lower. We therefore preprocess the data with a super-resolution network~\cite{Liang2021ICCV}, see supplementary.

\subsection{Image Synthesis}
Both our work and ~\cite{Dhariwal2021NEURIPS} use classifier networks to guide the generator. To ensure the models are not inadvertently optimizing for FID and IS, which also utilize a classifier network, we propose random-FID (rFID). For rFID, we calculate the Fr\'echet distance in the \texttt{pool\_3} layer of a randomly initialized inception network~\cite{Szegedy2015CVPR}.
The efficacy of random features for evaluating generative models has been demonstrated in~\cite{Naeem2020ICML}. Furthermore, we report sFID~\cite{Nash2021ICML} to assess spatial structure.
Lastly, sample fidelity and diversity are evaluated via precision and  recall~\cite{Kynknniemi2019NEURIPS}. 

In \tabref{tab:sotapr}, we compare StyleGAN-XL to the currently strongest GAN model (BigGAN-deep~\cite{Brock2019ICLR}) and diffusion models (CDM~\cite{Ho2022JMLR}, ADM~\cite{Dhariwal2021NEURIPS}) on ImageNet.
The values for ADM are calculated with and without additional methods (Upsampling \textbf{U} and Classifier Guidance \textbf{G}). For StyleGAN2, we report numbers by~\cite{Grigoryev2022ICLR}.
We find that StyleGAN-XL substantially outperforms all baselines across all resolutions in FID, sFID, rFID, and IS. 
An exception is recall, according
to which StyleGAN-XL’s sample diversity lies between BigGAN and
ADM, making progress in closing the gap between these model types.
BigGAN's sample quality is the best among all compared approaches, which comes at the price of significantly lower recall. 
StyleGAN-XL allows for the truncation trick to increase sample fidelity, i.e., we can interpolate a sampled style code $w$ with the class-wise mean style code $\bar{w}$.
We observe that for StyleGAN-XL, truncation does not increase precision, indicating that developing novel truncation methods for high-diversity GANs is an exciting research direction for future work.
Interestingly, StyleGAN-XL attains high diversity across all resolutions, which can be attributed to our progressive growing strategy. Furthermore, this strategy enables to scale to megapixel resolution successfully. Training at $1024^2$ for a single V100-day yields a noteworthy FID~of~$2.8$. At this resolution, we do not compare to baselines 
because of resource constraints as they are prohibitively expensive to train. 
visualizes generated samples at increasing resolutions.
\figref{fig:highres} visualizes generated samples at increasing resolutions. In the supplementary, we show additional  interpolations and qualitative comparisons to BigGAN and ADM.
\sotapr
\highres

\subsection{Inversion and Manipulation}
GAN-editing methods first \textit{invert} a given image into latent space, i.e., find a style code $w$ that reconstructs the image as faithful as possible when passed through $\bG_s$. Then, $w$ can be manipulated to achieve semantically meaningful edits~\cite{Goetschalckx2019ICCV,Shen2020TPAMI}.

\boldparagraph{Inversion.}
Standard approaches for inverting $\bG_s$ use either latent optimization~\cite{Abdal2019ICCV,Creswell2019NEURAL,Karras2020CVPR} or an encoder~\cite{Perarnau2016ARXIV,Alaluf2021ICCV,Tov2021TOG}. A common way to achieve low reconstruction error is to use an extended definition of the latent space: $\mathcal{W}+$. For $\mathcal{W}+$ a separate $\bw$ is chosen for each layer of $\bG_s$. However, as highlighted by~\cite{Zhu2020ECCV,Tov2021TOG}, this extended definition achieves higher reconstruction quality in exchange for lower editability. Therefore, ~\cite{Tov2021TOG} carefully design an encoder to maintain editability by mapping to regions of $\mathcal{W}+$ that are close to the original distribution of $\mathcal{W}$.
We follow~\cite{Karras2020CVPR} and use the original latent space $\mathcal{W}$.
We find that StyleGAN-XL already achieves satisfactory inversion results using basic latent optimization.
For inversion on the ImageNet validation set at $512^2$, StyleGAN-XL yields $\text{PSNR}=13.5$ on average, improving over BigGAN at $\text{PSNR}=10.8$. 
Besides better pixel-wise reconstruction, StyleGAN-XL's inversions are
semantically closer to the target images.
We measure the FID between reconstructions and targets, and StyleGAN-XL attains $\text{FID}=21.7$ while BigGAN reaches $\text{FID}=47.5$. 
For qualitative results, implementation details and additional metrics, we refer to the supplementary.

Given the results above, it is also possible to further refine the obtained reconstructions.~\cite{Roich2021ARXIV} recently introduced pivotal tuning inversion (PTI). PTI uses an initial inverted style code as a pivot point around which the generator is finetuned. Additional regularization prevents altering the generator output far from the pivot. Combining PTI with StyleGAN-XL allows us to invert both in-domain (ImageNet validation set) and out-of-domain images almost precisely. At the same time, the generator output remains perceptually smooth, see~\figref{fig:interpolations}.
\interpolations

\boldparagraph{Image Manipulation.}
Given the inverted images, we can leverage GAN-based editing methods~\cite{Voynov2020ICML,Haerkoenen2020NEURIPS,Shen2021CVPR,Kocasari2021WACV,Spingarn2021ICLR} to manipulate the style code $\bw$. In \figref{fig:editing}~(Left), we first invert a given source image via latent space optimization. 
We can then apply a manipulation directions obtained by, e.g., GANspace~\cite{Haerkoenen2020NEURIPS}.
Prior work~\cite{Jahanian2020ICLR} also investigates in-plane translation. This operation can be directly defined in the input grid of StyleGAN-XL. The input grid also allows performing extrapolation, see  \figref{fig:editing}~(Left).

An inherent property of StyleGAN is the ability of style mixing by supplying the style codes of two samples to different layers of $\bG_s$, generating a hybrid image. This hybrid takes on different semantic properties of both inputs. Style mixing is commonly employed for instances of a single domain, i.e., combining two human portraits. StyleGAN-XL inherits this ability and, to a certain extent, even generates out-of-domain combinations between different classes, akin to counterfactual images~\cite{Sauer2021ICLR}. This technique works best for aligned samples, similar to StyleGAN's originally favored setting, FFHQ. Curated examples are shown in \figref{fig:editing}~(Right).
\editing
\editingsupp

\section{Limitations and Future Work}
\label{sec:limitations}
Our contributions allow StyleGAN to accomplish state-of-the-art high-resolution image synthesis on ImageNet.  
Furthermore, applying it to big and small unimodal datasets is straightforward, and we also achieve state-of-the-art performance on FFHQ and Pokemon at resolution $1024^2$, see supplementary.
Exploring new editing methods and dataset generation~\cite{Chai2021CVPR,Li2022ARXIV} using StyleGAN-XL are exciting future avenues. Furthermore, future work may tackle an even larger megapixel dataset. However, a larger yet diverse dataset is not available so far. Current large-scale, high-resolution datasets are of single object classes or contain many similar images~\cite{Zhang2020ECCV,Fregin2018ICRA,Perot2020NEURIPS}. In the following, we discuss limitations of the current model, which should be addressed in the future.

\boldparagraph{Architectural Limitations.}
First, StyleGAN-XL is three times larger than StyleGAN3, constituting a higher computational overhead when used as a starting point for finetuning. Therefore, it will be worth exploring GAN distillation methods~\cite{Chang2020ACCV} that trade-off performance for model size. 
Second, we find StyleGAN3, and consequently, StyleGAN-XL, harder to edit, e.g., high-quality edits via $\mathcal{W}$ are noticeably easier to achieve with StyleGAN2. As already observed in~\cite{Karras2021NEURIPS}, StyleGAN3's semantic controllability is reduced for the sake of equivariance. 
However, techniques using the \textit{StyleSpace}~\cite{WU2021CVPRa}, e.g., StyleMC~\cite{Kocasari2021WACV}, tend to yield better results in our experiments, confirming the findings of concurrent work by~\cite{Alaluf2022ARXIV}. Furthermore, we remark that our framework can also easily be used with StyleGAN2 layers.

\clearpage
\begin{acks}
We acknowledge the financial support by the BMWi in the project KI Delta Learning (project number 19A19013O). 
Andreas Geiger was supported by the ERC Starting Grant LEGO-3D (850533). 
We would like to thank Kashyap Chitta, Michael Niemeyer, and Bo\v{z}idar Anti\'{c} for proofreading.
Lastly, we would like to thank Vanessa Sauer for her general support.
\end{acks}
\vspace{-20em}

\bibliographystyle{ACM-Reference-Format}
\bibliography{bibliography_long,bibliography_siggraph}

\appendix
In this supplemental document, we elaborate on increasing the resolution of ImageNet to one megapixel, compare to the baseline on a class containing humans, and specify the implementation details of our approach. The supplemental video shows additional samples and interpolations. We use the same mathematical notation as in the paper.

\section{Preprocessing ImageNet}
An initial challenge is the lack of high-resolution data; the mean resolution of ImageNet is $469\times387$.
Similar to the procedure used for generating CelebA-HQ\cite{Karras2018ICLR}, we preprocess the whole dataset with SwinIR-Large~\cite{Liang2021ICCV}, a recent model for real-world image super-resolution.
Of course, a trivial way of achieving good performance on this dataset would be to draw samples from a $256^2$ generative model and passing it through SwinIR. However, SwinIR adds significant computational overhead as it is $60$ times slower than our upsampling stack. Furthermore, this way, StyleGAN-XL's weights can be used for initialization when finetuning on other high-resolution datasets. Lastly, combining StyleGAN-XL and SwinIR would impair translation equivariance.

\section{Classes of unaligned Humans}
We observe that ADM~\cite{Dhariwal2021NEURIPS} generates more convincing human faces than StyleGAN-XL and BigGAN. Both GANs can synthesize realistic faces; however, the main challenge in this setting is that the dataset is unstructured, and the humans are not aligned.~\cite{Brock2019ICLR} remarked the particular challenge of classes containing details to which human observers are more sensitive. We show examples in~\figref{fig:aquamen}.
\aquamen

\section{Inference Speed}
GANs generate samples in a single forward pass, unlike diffusion models that must be applied several hundred or thousand times to generate a sample. \tabref{tab:speedcomparison} compares StyleGAN-XL to ADM. We find that StyleGAN-XL is several orders of magnitude faster. In defense of diffusion models, speeding up their sampling is an active area of research, and novel techniques~\cite{Watson2021ARXIV} may be able to reduce this gap in the future.
\speedcomparison

\section{Results on Unimodal Datasets}
StyleGAN-XL is designed to enable training on large and diverse datasets. However, applying it to big and small unimodal datasets is straightforward. 
In contrast to the configuration for ImageNet, we begin with ten layers at the lowest stage and add two layers per resolution stage. Furthermore, we do not employ classifier guidance.
\tabref{tab:extraresults} reports the results for both datasets at resolution $1024^2$, StyleGAN-XL achieves state-of-the-art performance on both.
\extraresults

\section{Additional Qualitative Results}

In the following, we present additional qualitative results. \figref{fig:interpsupp} shows additional interpolations between samples from different classes. \figref{fig:ffhqsamples} and \figref{fig:pokemonsamples} show samples on FFHQ $1024^2$ and Pokemon $1024^2$ respectively. Lastly, we compare BigGAN, ADM, and StyleGAN-XL on different ImageNet classes. For a  fair comparison, we do not use truncation or classifier guidance. Instead, we show images with the largest logits given by a VGG16 which corresponds to individual image quality.

\section{Implementation details}

\boldparagraph{Inversion.}
Following \cite{Karras2020CVPR}, we use basic latent optimization in $\mathcal{W}$ for inversion. Given a target image, we first compute its average style code $\bar{\bw}$ by running $10000$ random latent codes $\bz$ and target specific class samples $\bc$ through the mapping network. As the class label of the target image is unknown, we pass it to a pretrained classifier. We then use the classifier logits as a multinomial distribution to sample $\bc$. In our experiments, we use Deit-base~\cite{Touvron2021ICML} as a classifier, but other choices are possible.
At the beginning of optimization , we initialize $\bw = \bar{\bw}$. The components of $\bw$ are the only trainable parameters. The optimization runs for 1000 iterations using the Adam optimizer~\cite{Kingma2015ICLR} with default parameters. 
We optimize the LPIPS~\cite{Zhang2018CVPR} distance between the target image and the generated image.
For StyleGAN-XL, the maximum learning rate is $\lambda_{max} = 0.05$. It is ramped up from zero linearly during the first 50 iterations and ramped down to zero using a cosine schedule during the last 250 iterations. 
For BigGAN, we empirically found $\lambda_{max} = 0.001$ and a ramp-down over the last 750 iterations to yield the best results. 
All inversion experiments are performed at resolution $512^2$ and computed on $5k$ images ($10$\% of the validation set). We report the results in \tabref{tab:invresults} and show qualitative results in \figref{fig:inversion}.
\invresults

\boldparagraph{Training StyleGAN3 on ImageNet.}
For training StyleGAN3, we use the official PyTorch implementation\footnote{\url{https://github.com/NVlabs/stylegan3.git}}.
The results in 
\figref{fig:teaser} 
are computed with the StyleGAN3-R configuration on resolution $256^2$ until the discriminator has seen $10$ million images. We find that StyleGAN3-R and StyleGAN3-T converge to similar FID without any changes to their training paradigm. The run with the best FID score was selected from three runs with different random seeds. We use a channel base of $16384$ and train on $8$ GPUs with total batch size $256$, $\gamma=0.256$. The remaining settings are chosen according to the default configuration of the code release.
For the ablation study in  
\tabref{tab:ablation}
, we use the StyleGAN3-T configuration as baseline since StyleGAN-XL builds upon the translational-equivariant layers of StyleGAN3.
We train on $4$ GPUs with total batch size $256$ and batch size $32$ per GPU, $\gamma=0.25$, and disable augmentation.

\boldparagraph{Training \& Evaluation.}
For all our training runs, we do not use data amplification via \textit{x-flips} following~\cite{Karras2020CVPR}. Furthermore, we evaluate all metrics using the official StyleGAN3 codebase. For the baseline values in
\tabref{tab:sotapr}
we report the numbers of~\cite{Dhariwal2021NEURIPS}. The official codebase of ADM\footnote{\url{https://github.com/openai/guided-diffusion}} provides files containing $50$k samples for ADM and BigGAN. We utilize the provided samples to compute rFID. Following ~\cite{Dhariwal2021NEURIPS}, we compute precision and recall between $10$k real samples and $50$k generated samples. \tabref{tab:imagenetlowres} reports the results on ImageNet at lower resolutions.

\imagenetlowres

\boldparagraph{Layer configurations.}
We start progressive growing at resolution $16^2$ using $11$ layers. The layer specifications are computed according to \cite{Karras2021NEURIPS} and remain fixed for the remaining training. For the next stage, at resolution $32^2$, we discard the last $2$ layers and add $7$ new ones. The specifications for the new layers are computed according to \cite{Karras2021NEURIPS} for a model with resolution $32^2$ and $16$ layers. Continuing this strategy up to resolution $1024^2$ yields the flexible layer specification of StyleGAN-XL in \figref{fig:layerspecs}.

\interpsupp
\inversion
\ffhqsamples
\pokemonsamples
\samplesa
\samplesb
\samplesc
\layerspecs

\end{document}